\documentclass[10pt,twocolumn,letterpaper]{article}

\usepackage{cvpr}              % To produce the CAMERA-READY version
\usepackage[dvipsnames]{xcolor}

\usepackage[noend]{algpseudocode}
\usepackage{multirow}
\usepackage{booktabs}
\usepackage{array}
\usepackage{amsmath}
\usepackage{diagbox}
\usepackage{xspace}
\usepackage{bm}
\usepackage[ruled,linesnumbered,vlined]{algorithm2e}
\usepackage{graphicx}
\usepackage{subcaption}
\usepackage{wrapfig}
\usepackage{makecell}
\usepackage{ragged2e}
\usepackage{xcolor}
\newcommand{\bx}{\bm{x}}

\newcommand{\btheta}{\bm{\theta}}

\newcommand{\mx}{\mathcal{X}}

\newcommand{\mt}{\mathcal{T}}
\newcommand{\ms}{\mathcal{S}}
\newcommand{\ml}{\mathcal{L}}

\makeatletter
\DeclareRobustCommand\onedot{\futurelet\@let@token\@onedot}
\def\@onedot{\ifx\@let@token.\else.\null\fi\xspace}

\renewcommand{\paragraph}{%
	\@startsection{paragraph}{4}{\z@}%
	{0.1em \@plus 0.5ex \@minus 0.2ex}{-1em}%
	{\normalsize\bf}%
}
\makeatother

\usepackage{amsmath,amsfonts,bm,mathtools,bbm}

\def\eqref#1{equation~\ref{#1}}
\def\1{\bm{1}}

\def\vx{{\bm{x}}}

\DeclareMathAlphabet{\mathsfit}{\encodingdefault}{\sfdefault}{m}{sl}
\SetMathAlphabet{\mathsfit}{bold}{\encodingdefault}{\sfdefault}{bx}{n}

\definecolor{cvprblue}{rgb}{0.21,0.49,0.74}
\usepackage[pagebackref,breaklinks,colorlinks,citecolor=cvprblue]{hyperref}

\def\paperID{DD Workshop} % *** Enter the Paper ID here
\def\confName{CVPR}
\def\confYear{2024}

\title{Large-scale Dataset Pruning with Dynamic Uncertainty}

\author{
Muyang He$^{1,2}$\thanks{Intern at Beijing Academy of Artificial Intelligence} \quad Shuo Yang$^3$ \quad Tiejun Huang$^{1,2}$ \quad Bo Zhao$^1$\thanks{Corresponding author} \\
$^1$Beijing Academy of Artificial Intelligence\\
$^2$Peking University \quad
$^3$University of Technology Sydney \\
{\tt \small isaache@pku.edu.cn, zhaobo@baai.ac.cn}
}

\begin{document}
\maketitle
\begin{abstract}
 The state of the art of many learning tasks, e.g., image classification, is advanced by collecting larger datasets and then training larger models on them. As the outcome, the increasing computational cost is becoming unaffordable. In this paper, we investigate how to prune the large-scale datasets, and thus produce an informative subset for training sophisticated deep models with negligible performance drop. We propose a simple yet effective dataset pruning method by exploring both the prediction uncertainty and training dynamics. We study dataset pruning by measuring the variation of predictions during the whole training process on large-scale datasets, i.e., ImageNet-1K and ImageNet-21K, and advanced models, i.e., Swin Transformer and ConvNeXt. Extensive experimental results indicate that our method outperforms the state of the art and achieves $25\%$ lossless pruning ratio on both ImageNet-1K and ImageNet-21K. The code and pruned datasets are available at \url{https://github.com/BAAI-DCAI/Dataset-Pruning}.
\end{abstract}    
\section{Introduction}

The emerging large models, e.g., vision transformers \cite{dosovitskiy2021an, liu2021swin, fang2023eva} in computer vision, can significantly outperform the traditional neural networks, e.g., ResNet \cite{he2016deep}, when trained on large-scale datasets, e.g., ImageNet-21K \cite{deng2009imagenet} and JFT-300M \cite{sun2017revisiting}. However, storing large datasets and training on them are expensive and even unaffordable, which prevents individuals or institutes with limited resources from exceeding the state of the art. How to improve the data-efficiency of deep learning and achieve good performance with less training cost is a long-standing problem. 

It is known that large-scale datasets have many redundant and less-informative samples which contribute little to model training. {Dataset pruning (or coreset selection) \cite{chen2010super, sener2018active, gal2017deep, shen2018deep, toneva2018an, feldman2020introduction, sorscher2022beyond, guo2022deepcore, xia2023moderate, yang2023dataset, coleman2020selection, paul2021deep, zheng2023coveragecentric} }aims to remove those less-informative training samples and remain the informative ones of original dataset, such that models trained on the remaining subset can achieve comparable performance. Nevertheless, how to find the informative subset is a challenging problem, especially for large-scale datasets, since sample selection is a combinatorial optimization problem. 

\paragraph{Limitations of previous work.} {Existing dataset pruning works 
% \cite{chen2010super, sener2018active, gal2017deep, shen2018deep, toneva2018an, sorscher2022beyond, xia2023moderate, yang2023dataset} 
select important samples based on geometry/distribution \cite{chen2010super, sener2018active, xia2023moderate, zheng2023coveragecentric}, prediction uncertainty \cite{shen2018deep,coleman2020selection,gal2017deep}, training error/gradient \cite{paul2021deep, toneva2018an}, optimization \cite{borsos2020coresets, killamsetty2021retrieve, killamsetty2021glister,yang2023dataset}, etc.} However, these methods have several major limitations: 1) Confusion between hard samples and mislabeled samples. Both of them lead to more prediction errors and larger loss. Exploiting the uncertainty or prediction error solely may obtain inferior subsets. 2) Ignoring training dynamics. Existing methods generally utilize a pretrained model to select samples. The geometry of data distribution thus does not reflect that throughout the training process. 3) Difficulty in scalability and generalization. Most of pruning methods focus on small-scale datasets, e.g., CIFAR-10/100 \cite{CIFAR10} which contains only $50$K training images with $32\times32$ resolution, and traditional convolutional models, e.g., ResNet. They have difficulty in scaling up to large datasets with millions of samples due to the high computational complexity and memory cost. 

% Different datasets have varying properties of data distribution and redundancy. Different architectures also have varying data demand. Besides, some methods with high computational complexity, especially those optimization-based methods, are feasible in the small-scale setting, but they are not scalable to datasets with millions of samples due to intolerable time or resource costing.

\begin{figure*}[th]
		\centering
     \vspace{-10pt}
	\begin{minipage}[c]{0.42\textwidth}
		\centering
		\includegraphics[width=\textwidth]{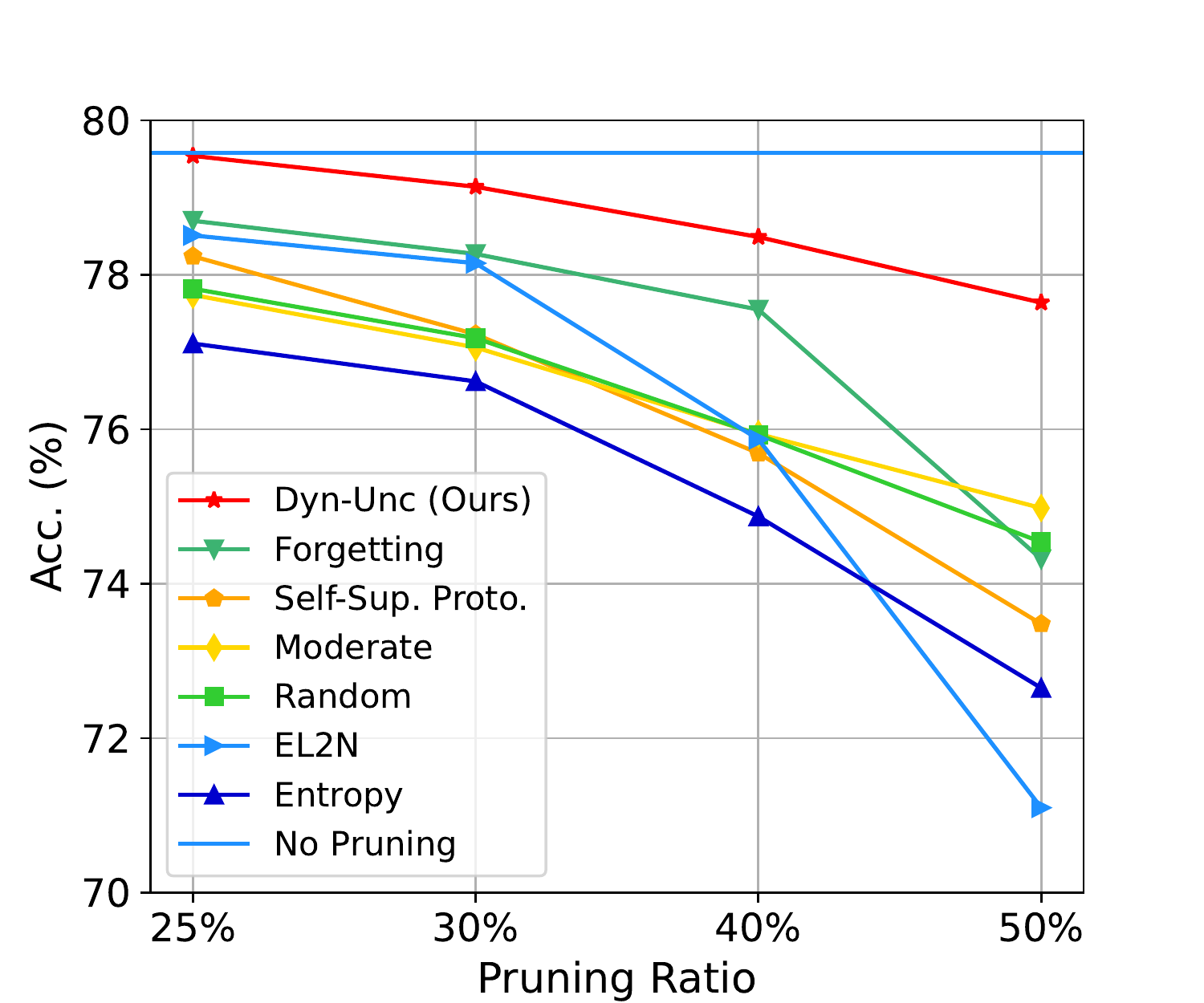}
		\subcaption{Top-1 accuracy on ImageNet-1K-val}
		\label{fig:1k_res_val}
	\end{minipage}
	\begin{minipage}[c]{0.42\textwidth}
		\centering
		\includegraphics[width=\textwidth]{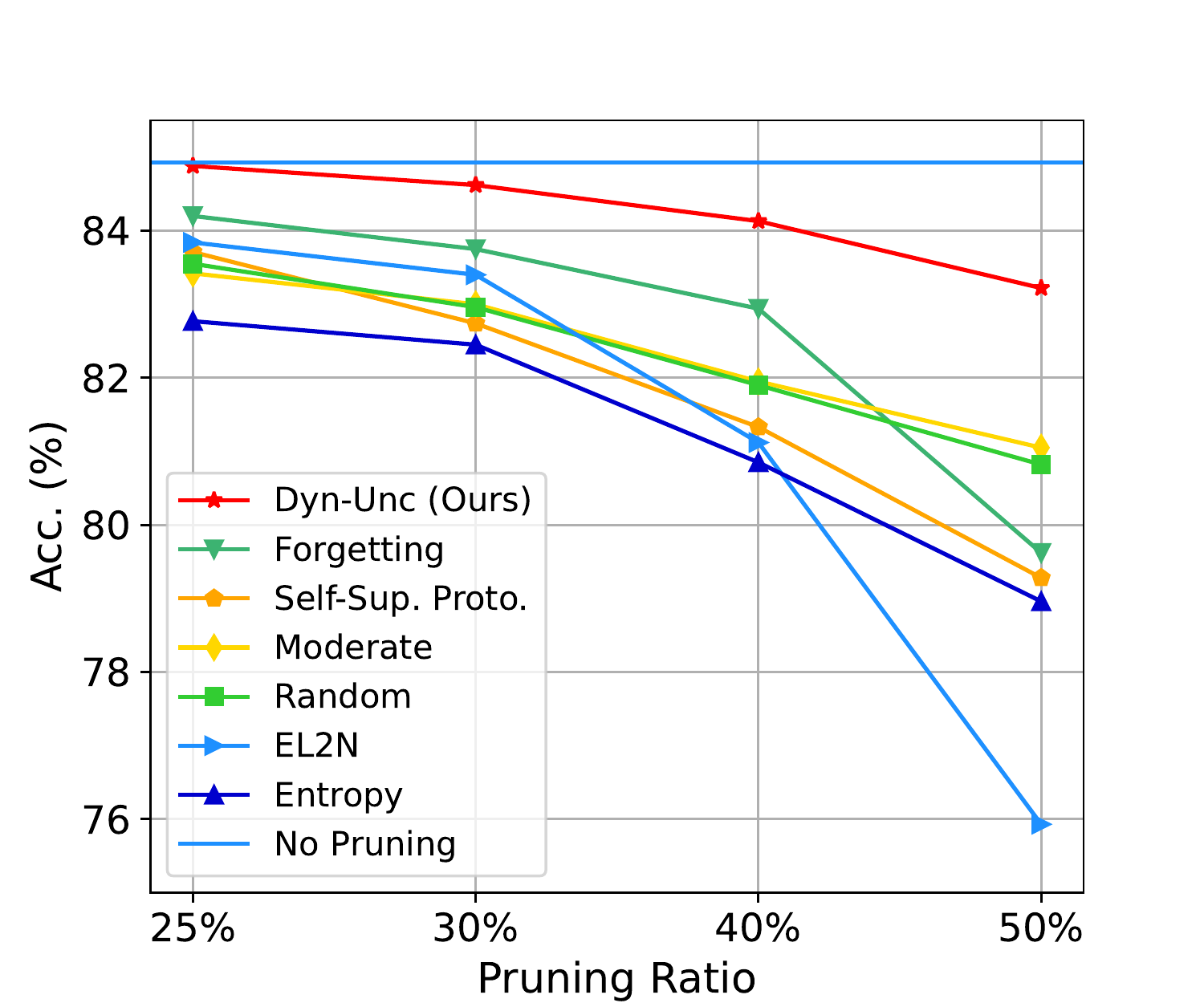}
		\subcaption{Top-1 accuracy on ImageNet-1K-ReaL}
		\label{fig:1k_res_ReaL}
	\end{minipage} 
	\caption{{Comparison to state-of-the-art dataset pruning methods on ImageNet-1K.}}
	\label{fig:1k_res}
     \vspace{-5pt}
\end{figure*}

\paragraph{Our method.} In this paper, we design a simple yet effective sample selection metric dubbed Dynamic-Uncertainty (\emph{Dyn-Unc}) for dataset pruning. We extract the prediction uncertainty by measuring the variation of predictions in a sliding window, and then introduce the training dynamics into it by averaging the variation throughout the whole training process. To address the first limitation, \emph{Dyn-Unc} favors the uncertain samples rather than easy-to-learn ones or hard-to-learn ones during model training. \emph{Dyn-Unc} can better separate informative samples from noise or mislabeled samples, as the model will gradually produce less uncertainty on mislabeled samples. For the second limitation, \emph{Dyn-Unc} introduces the training dynamics and learns the property of data throughout the training process instead of a certain stage. 
Since our method requires only one-run model training on the whole dataset before pruning it, it is scalable to large datasets.

% Experimental results demonstrate that \emph{Dyn-Unc} achieves the best results on ImageNet-1K and ImageNet-21K and good generalization performance to unseen architectures. 
As shown in \cref{fig:1k_res} and \cref{tab:21k_res}, our \emph{Dyn-Unc} significantly surpasses the previous state-of-the-art methods and achieve $25$\% lossless pruning ratio on ImageNet-1K and ImageNet-21K. Besides, experimental results (\cref{tab:cross_res}, \cref{tab:cross_res_compa}) demonstrate that the coreset selected by \emph{Dyn-Unc} using one architecture can generalize well to other unseen architectures.

\section{Related work}

\subsection{Dataset Pruning}
Dataset pruning is mainly rooted in coreset selection, which focuses on selecting a small but highly informative subset from a large dataset for training models.
Dataset pruning methods typically calculate a scalar score for each training example based on predefined criteria. These can be grouped into geometry-based, uncertainty-based, error-based, bilevel-optimization-based, etc. For instance, (1) \emph{Geometry}-based methods employ geometric measures in the feature space to select samples that preserve original geometry. A representative method is \emph{Herding} \cite{chen2010super}, which incrementally selects a sample that enables the new subset center to be closest to the whole-dataset center. \emph{K-Center} \cite{sener2018active, wolf2011kcenter} method minimizes the largest distance from any data point to its nearest center in the selected subset, while \emph{Moderate} \cite{xia2023moderate} targets at data points with distances close to the median. (2) \emph{Uncertainty}-based methods identify the most challenging samples, which are typically defined as instances that the model exhibits the least confidence in \cite{shen2018deep,coleman2020selection,gal2017deep}, or those that lie near the decision boundary where prediction variability is high \cite{ducoffe2018adversarial,margatina2021active}. For example, \cite{chang2017active} adjust the weights assigned to training samples by taking into account the variance of their predictive distribution. (3) \emph{Error}-based methods focus on the samples that contribute most to the loss. \emph{GraNd} and \emph{EL2N} \cite{paul2021deep} methods use the gradient norm and the L2-distance between the normalized error of predicted probabilities and one-hot labels to identify important samples. Another popular method, \emph{Forgetting} \cite{toneva2018an}, considers retaining the most forgettable samples. These are defined as those ones flipping most frequently from being correctly classified to being misclassified throughout training process. (4) There also exist some methods based on \emph{bilevel optimization} \cite{borsos2020coresets, killamsetty2021retrieve, killamsetty2021glister}, which optimize the subset selection (weights) that leads to best-performing models trained on the subset. However, these methods have difficulty in scaling up to large datasets due to the complex bilevel optimization. 
% Other methods in this category examine a score function based on each example's relative contribution to the total loss across all training samples \cite{bachem2015coresets, munteanu2018coresets}.

While the above methods exhibit decent performance on small datasets and traditional (convolutional) models, we have empirically discovered that those studies based on small datasets have difficulty in generalizing to larger datasets and modern large models (e.g., Vision Transformers). Recognizing this limitation, our work takes an unprecedented step towards pruning highly extensive datasets, specifically ImageNet-1K and ImageNet-21K, and evaluating them by using state-of-the-art large models like Swin Transformer and ConvNeXt. This advancement marks a crucial milestone in the field, addressing the need for effective data pruning methods that can handle very large datasets and cutting-edge models.

\subsection{Dataset Distillation}
Another research topic that focuses on reducing the amount of training data is dataset distillation \cite{wang2018dataset,zhao2021DC,zhao2023data}. Instead of selecting a subset, these studies learn to \emph{synthesize} a highly compact yet informative dataset that could be utilized for model training from scratch. The synthesis process involves aligning various factors such as performance \cite{wang2018dataset,such2020generative,nguyen2021dataset1}, gradient \cite{zhao2021DC,zhao2021DSA}, distribution \cite{zhao2023DM,zhao2022synthesizing}, feature map \cite{wang2022cafe}, and training trajectories \cite{cazenavette2022distillation,cui2023scaling}, between the model trained on the original dataset and the one trained on the synthetic data. Despite their promising prospects, these techniques, in their current state, cannot scale to large-scale datasets and models due to the computational load associated with the pixel-level optimization for image synthesis \cite{cui2023scaling}. In addition, existing dataset distillation methods show marginal performance improvements over subset selection methods in the experiments of synthesizing a large number of  training samples \cite{zhao2023DM}, since optimizing increasingly more variables (pixels) is challenging. 

\section{Method}

\subsection{Notation}
Given a large-scale labeled dataset $\mathcal{T}$ containing $n$ training samples, denoted by $\mathcal{T}=\{(\vx_1, y_1), (\vx_2, y_2),\dots,(\vx_n,y_n)\}$, where $\bx\in\mx\subset \mathbb{R}^d$ is the data point and $y\in\{0,\dots,C-1\}$ is the label belonging to $C$ classes. 
The goal of dataset pruning is to remove the less-informative training samples and then output a subset $\mathcal{S}\subset\mathcal{T}$ with $|\mathcal{S}| < |\mathcal{T}|$, namely \emph{coreset}, which can be used to train models and achieve comparable generalization performance to those trained on the whole dataset:  
\begin{equation}
|\mathbb{E}_{\bx \sim P_{\mathcal{D}}}[\ell(\phi_{\btheta^\mt}(\bx),y)] - \mathbb{E}_{\bx \sim P_{\mathcal{D}}}[\ell(\phi_{\btheta^\ms}(\bx),y)]| < \epsilon.
\end{equation} 
$P_{\mathcal{D}}$ is the real data distribution. $\phi_{\btheta^\mt}(\cdot)$ and $\phi_{\btheta^\ms}(\cdot)$ are models parameterized with $\btheta$ and trained on $\mt$ and $\ms$ respectively. $\epsilon$ is a small positive number. The pruning ratio is calculated as $r = 1-\frac{|\mathcal{S}|}{|\mathcal{T}|}$. 
% We follow the notations in \cite{zhao2021DC}.

\subsection{Dataset Pruning}
It is a common practice to classify training samples into three categories \cite{chang2017active, swayamdipta2020dataset}: easy, uncertain and hard samples. Previous works have shown that selecting easy, uncertain or hard samples may work well in different learning tasks. For example, \emph{Herding} \cite{chen2010super} method tends to select easy samples that are close to class center. \emph{GraNd} and \emph{EL2N} \cite{paul2021deep} pick hard samples with large training gradients or errors. Different from them, \cite{chang2017active, swayamdipta2020dataset} measure the uncertainty of training samples and keep those with large prediction uncertainty. 
Thus, the best selection strategy is determined by multiple factors, including dataset size, data distribution, model architecture and training strategy. 

% Inspired by \cite{swayamdipta2020dataset}, \cite{chang2017active},
% we use the \textbf{prediction variance} to measure the essential
% of each training sample. In terms of the classification task, the last layer of model is the classifier which outputs the probability
% assigned by model of sample belonging to that class. And a changing
% probability means the sample is with higher uncertainty and ambiguity and confuses model from time to time. Institutionally,
% these samples are hard. Following former works, we believe harder samples are more informative, and thus we should keep them as the coreset. At the same time, \cite{swayamdipta2020dataset}, \cite{chang2017active} shows that samples with higher variance contributes more to the generalization ability. Therefore, our method is uncertainty-based and then variance-based.

\paragraph{Prediction Uncertainty.} We design a simple yet effective large-scale dataset pruning method that selects samples by considering both prediction uncertainty and training dynamics. Larger dataset probably contains more easy samples that are replaceable for model training and hard/noisy samples that cannot be learned due to inconsistency between images and labels \cite{ridnik2021imagenetk}. In other words, the model trained on the rest can generalize well to those easy samples, meanwhile overfitting noisy samples does not improve model's generalization performance. Hence, we choose to select samples with large prediction uncertainty and remove the rest. 

% We define the uncertainty based on the variance of model prediction. 
% Formally, 
Given a model $\btheta^k$ trained on the whole dataset at $k_\text{th}$ epoch, its prediction on the target label is denoted as $\mathbb{P}(y|\bm{x}, \btheta^k)$. The uncertainty is defined as the standard deviation of the predictions in successive $J$ training epochs with models $\{\btheta^k, \btheta^{k+1}, ...,\btheta^{k+J-1}\}$:
\begin{equation}
\begin{aligned}
\label{eq:Uk}
    U_k(\bm{x}) = \sqrt{\frac{\Sigma_{j=0}^{J-1} [\mathbb{P}(y|\bm{x}, \btheta^{k+j}) - \bar{\mathbb{P}}]^2}{J-1}}, \;\;\; \\
    \text{where} \;\;\; \bar{\mathbb{P}} = \frac{\Sigma_{j=0}^{J-1} \mathbb{P}(y|\bm{x}, \btheta^{k+j})}{J}.
\end{aligned}
\end{equation}
$J$ is the range for calculating the deviation. 

% during the training process, the model predicts a probability vector of all classes of label for each sample, and $\mathbb{P}(y_i|\bm{x}_i)$ is the predicted probability of target label. At the current iteration $t$, we consider the previous $w$ predictions and denote the variance $\mathbb{VAR}[\mathbb{P}_{t-w+1}(y_i|\bm{x}_i),\mathbb{P}_{t-w+2}(y_i|\bm{x}_i),\dots,\mathbb{P}_t(y_i|\bm{x}_i)]$ as an observation of prediction variance.

\paragraph{Training Dynamics.} Various uncertainty metrics have been designed in previous works, which are typically measured in a certain epoch, for example, in the last training epoch. However, the uncertainty metric may vary for models at different training epochs. Hence, we enhance the uncertainty metric with training dynamics. Specifically, we measure uncertainty in a sliding window with the length $J$, and then average the uncertainty throughout the whole training process:
\begin{equation}
\label{eq:dynunc}
    U(\bm{x}) = \frac{\Sigma_{k=0}^{K-J-1} U_k(\bm{x})}{K-J} ,
\end{equation}
where $K$ is the training epochs.

The training algorithm is depicted in \cref{algo:dyn unc}. Given a dataset $\mt$, we train the model $\phi_{\btheta}$ to compute the dynamic uncertainty of each sample. In each training iteration, the prediction and loss are computed for every sample. Note that the computation can be in parallel with GPU. The loss averaged on a batch is used to update model parameters. Meanwhile, we compute the uncertainty of each sample based on its predictions over the past $J$ epochs using \cref{eq:Uk}. After $K$ training epochs, the dynamic uncertainty is calculated using \cref{eq:dynunc}. Then, we sort all samples in the descending order of dynamic uncertainty $U(\cdot)$ and output the front $(1-r)\times|\mt|$ samples as the pruned dataset $\ms$, where $r$ is the pruning ratio.

% Taking advantage of the knowledge implied by the whole training process, especially the variety between iterations, variance of the whole process rather than an single observation will be taken into consideration. Through an $N$ epochs' training process, there are $N-w$ observations of prediction variance for one sample (trained model excluded). And we simply
% average them as a score to evaluate a sample, called \textbf{variance-average score}. 
% \begin{equation*}
%     \text{variance-average score}=\frac{\sum_{t=w}^{N-1}\mathbb{VAR}[\mathbb{P}_{t-w+1}(y_i|\bm{x}_i),\mathbb{P}_{t-w+2}(y_i|\bm{x}_i),\dots,\mathbb{P}_t(y_i|\bm{x}_i)]}{N-w}
% \end{equation*}

% Then, given the compression rate $r$, we keep the $r|\mathcal{T}|$
% samples with highest variance-average-score as the coreset.
% In practice, we use the standard deviation to simplify calculation.
\begin{algorithm}%[H]
% \small
\KwIn{Training set $\mt$, pruning ratio $r$.}
% \justifying{
\textbf{Required}: Deep model $\phi_{\btheta}$ parameterized with $\btheta$, data augmentation $\mathcal{A}$, training epochs $K$, length of uncertainty window $J$, learning rate $\eta$.

\For{$k = 0, \cdots, K-1$} {
        $\triangleright$ In parallel on GPUs.
        
	Sample a batch $\mathcal{B}\sim\mt$. 
 
        \For{$(\bm{x}_i,y_i) \in \mathcal{B}$}{ 
        Compute prediction $\mathbb{P}(y_i|\bm{x}_i, \btheta)$  and loss $\ell(\phi_{\btheta}(\mathcal{A}(\bx_i)),y_i)$
        
        \If{$k\geq J$}{
        	Compute uncertainty $U_{k-J}(\bm{x}_i)$ using \cref{eq:Uk} 
        }
        }

    Update $\btheta\leftarrow \btheta - \eta\nabla_{\btheta}\ml$, where $\ml = \frac{\Sigma\ell(\phi_{\btheta}(\mathcal{A}(\bx_i)),y_i)}{|B|}$
     
} 

\For{$(\bm{x}_i,y_i) \in \mt$}{
    Compute dynamic uncertainty $U(\bm{x_i})$ using \cref{eq:dynunc}
}

Sort $\mt$ in the descending order of $U(\cdot)$

$\ms \leftarrow$ front $(1-r)\times|\mt|$ samples in the sorted $\mt$

\KwOut{Pruned dataset $\ms$}

\caption{ \emph{Dyn-Unc}.}
\label{algo:dyn unc}
\end{algorithm}

\begin{figure*}
\begin{minipage}[c]{0.92\textwidth}
	\begin{minipage}[c]{0.24\textwidth}
		\centering
		\includegraphics[width=\textwidth]{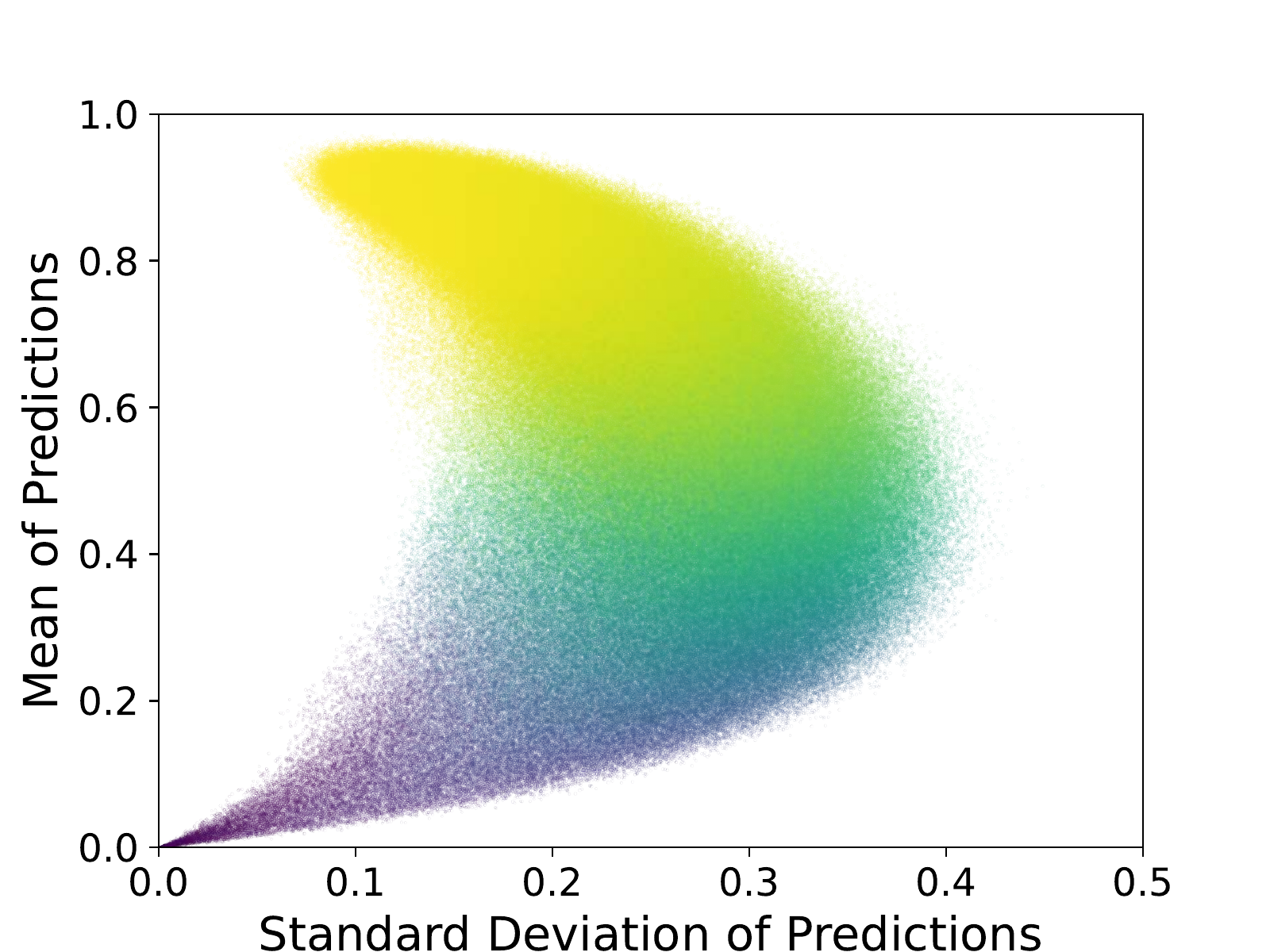}
		\subcaption{ImageNet-1K Train Set}
		\label{fig:dm_total}
	\end{minipage}
	\begin{minipage}[c]{0.24\textwidth}
		\centering
		\includegraphics[width=\textwidth]{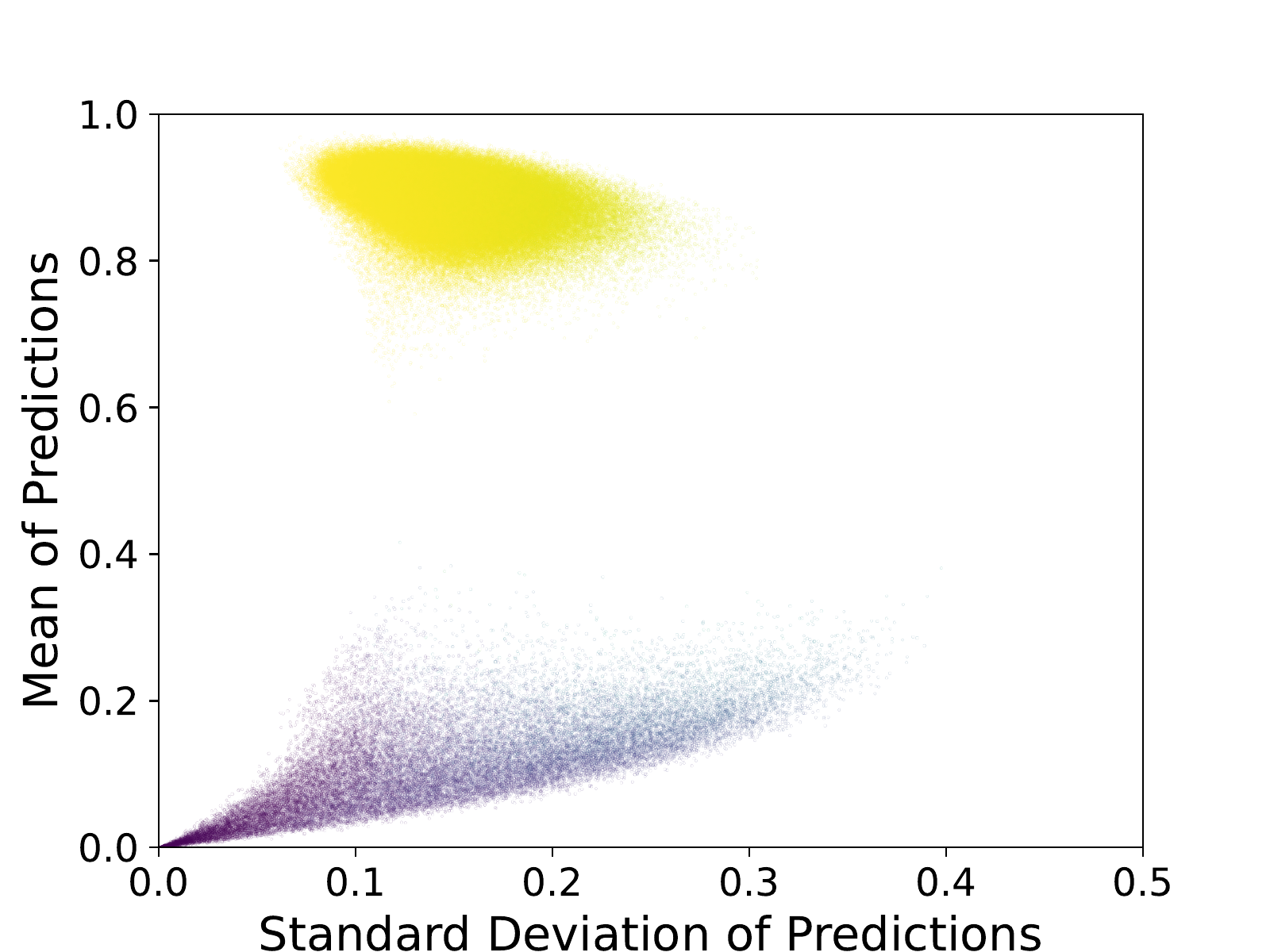}
		\subcaption{Dyn-Unc (Ours) Pruned}
		\label{fig:dm_75_ours}
	\end{minipage}
        \begin{minipage}[c]{0.24\textwidth}
		\centering
		\includegraphics[width=\textwidth]{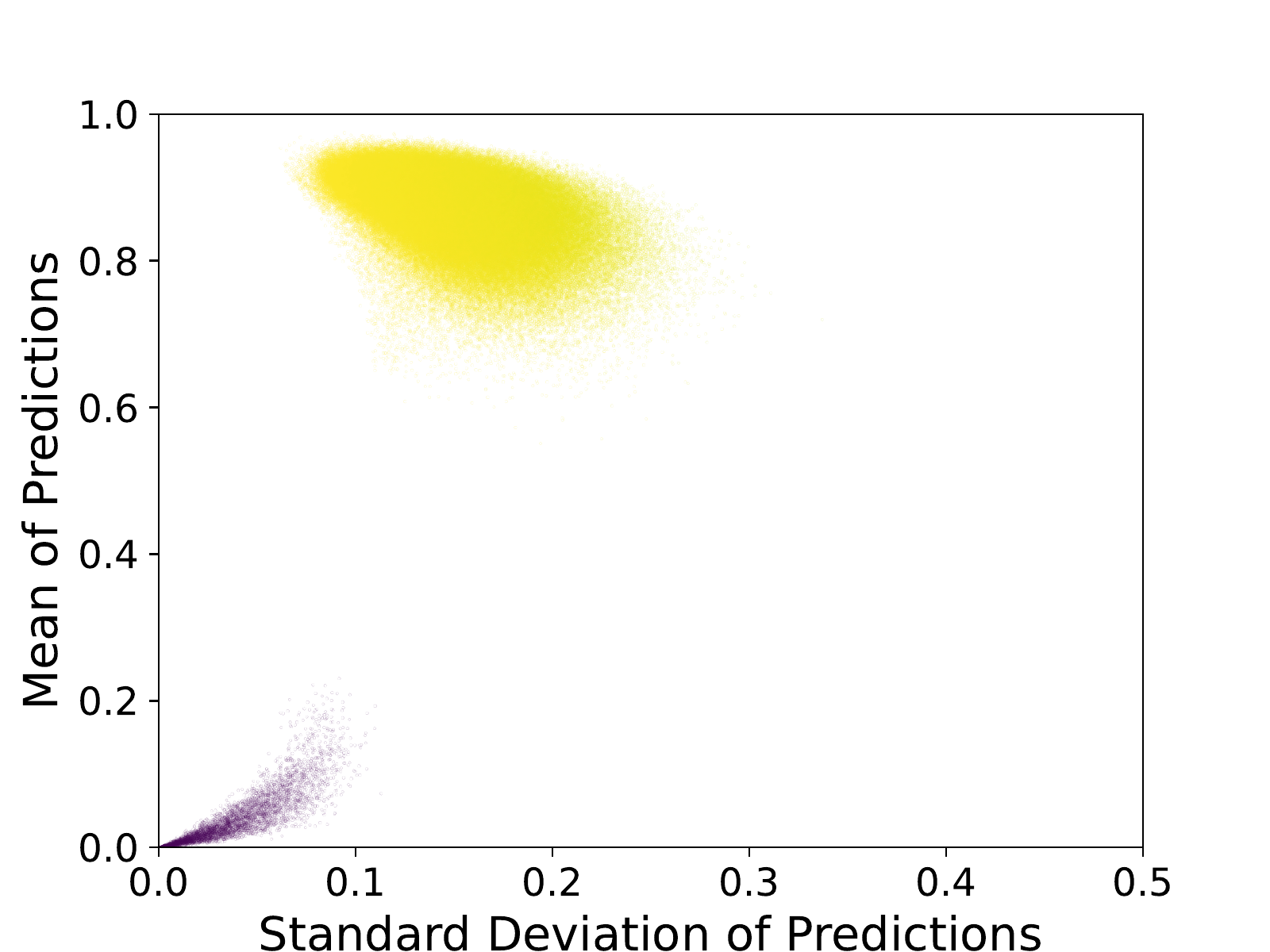}
		\subcaption{Forgetting Pruned}
		\label{fig:dm_75_forgetting}
	\end{minipage}
        \begin{minipage}[c]{0.24\textwidth}
		\centering
		\includegraphics[width=\textwidth]{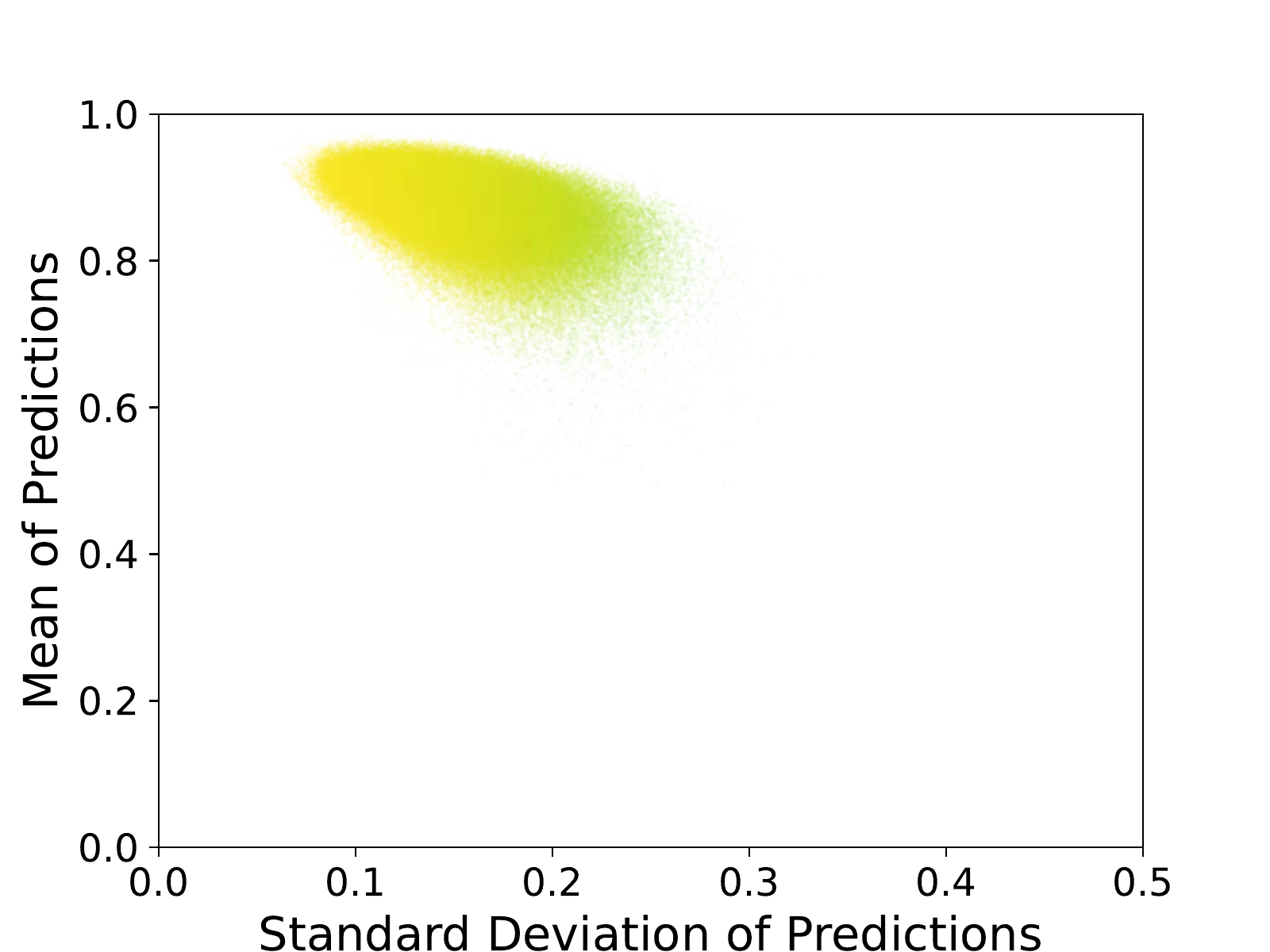}
		\subcaption{EL2N Pruned}
		\label{fig:dm_75_el2n}
	\end{minipage}\\
        \begin{minipage}[c]{0.24\textwidth}
		\centering
		\includegraphics[width=\textwidth]{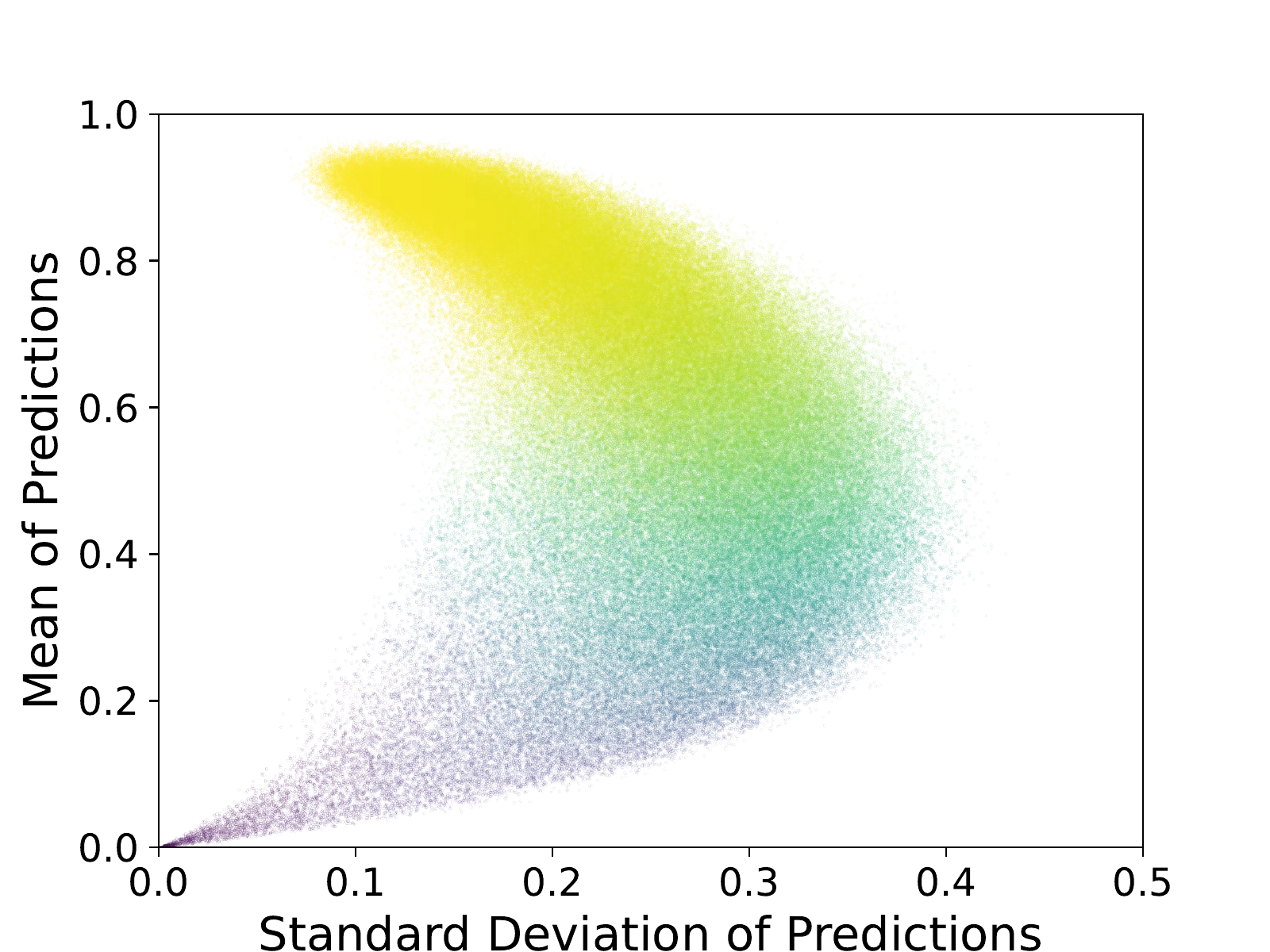}
		\subcaption{Random Pruned}
		\label{fig:dm_75_rand}
	\end{minipage}
 	\begin{minipage}[c]{0.24\textwidth}
		\centering
		\includegraphics[width=\textwidth]{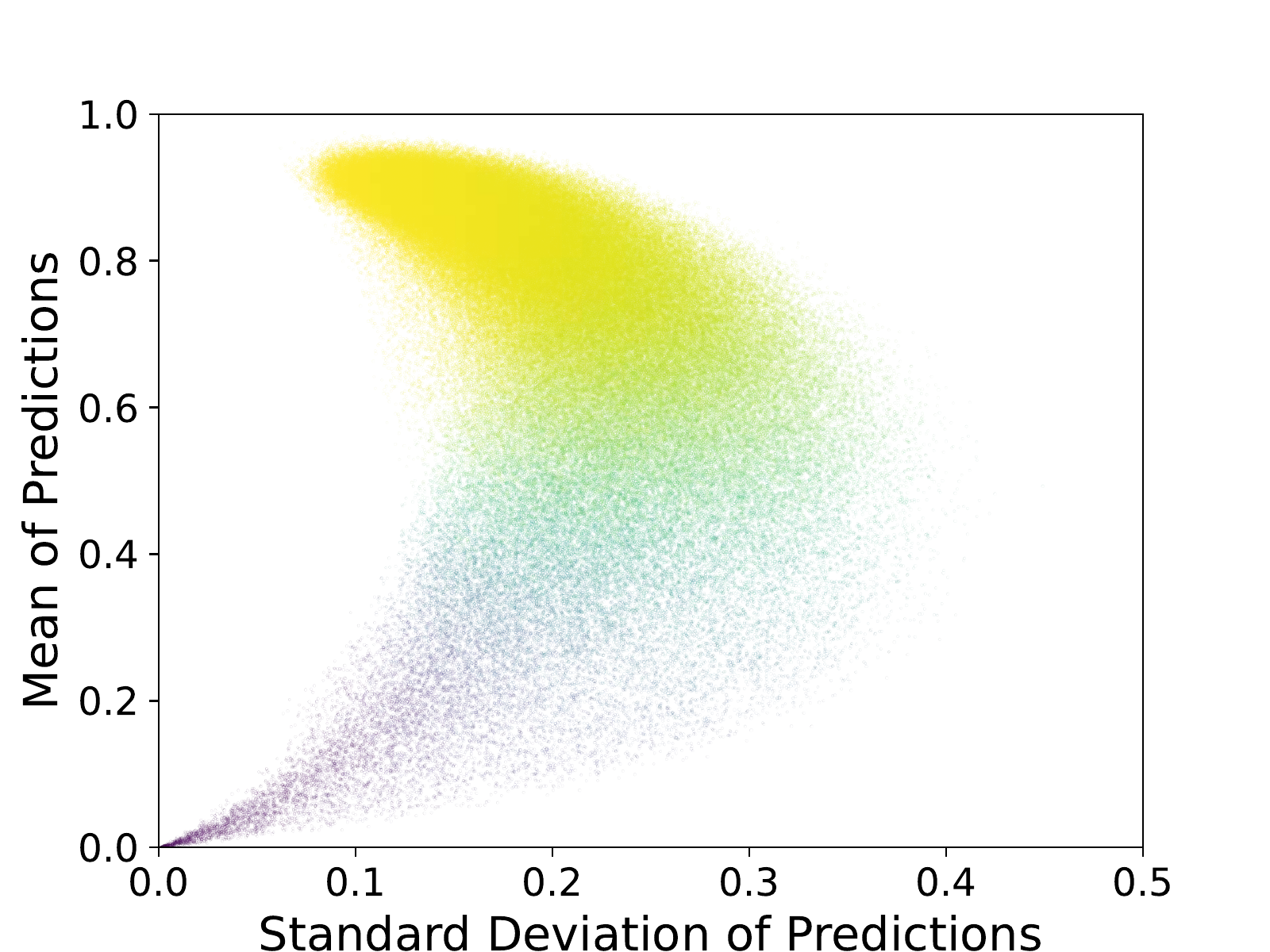}
		\subcaption{Self-Sup. Proto. Pruned}
		\label{fig:dm_75_ssp}
	\end{minipage}
	\begin{minipage}[c]{0.24\textwidth}
		\centering
		\includegraphics[width=\textwidth]{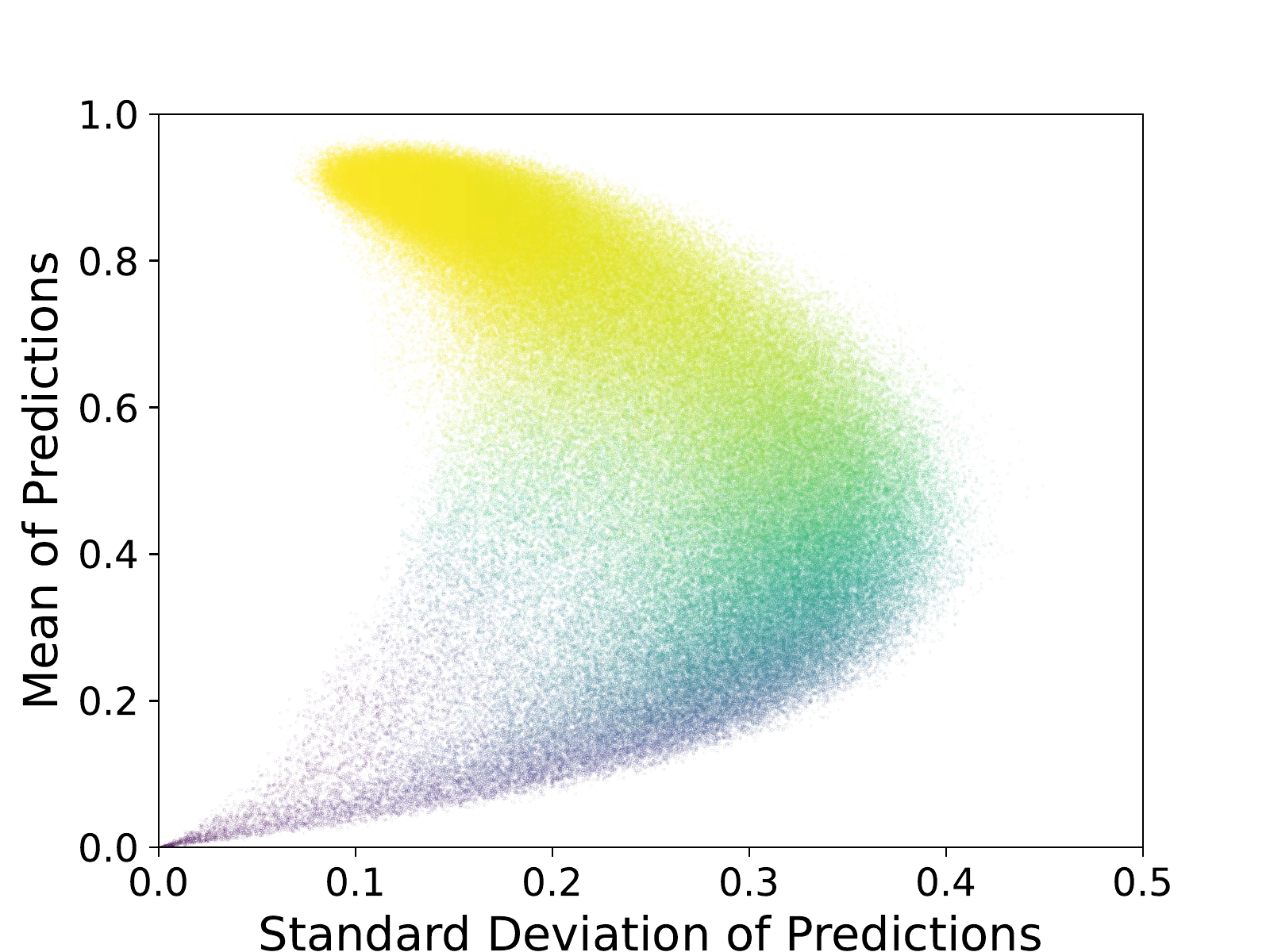}
		\subcaption{Moderate Pruned}
		\label{fig:dm_75_mds}
	\end{minipage}
        \begin{minipage}[c]{0.24\textwidth}
		\centering
		\includegraphics[width=\textwidth]{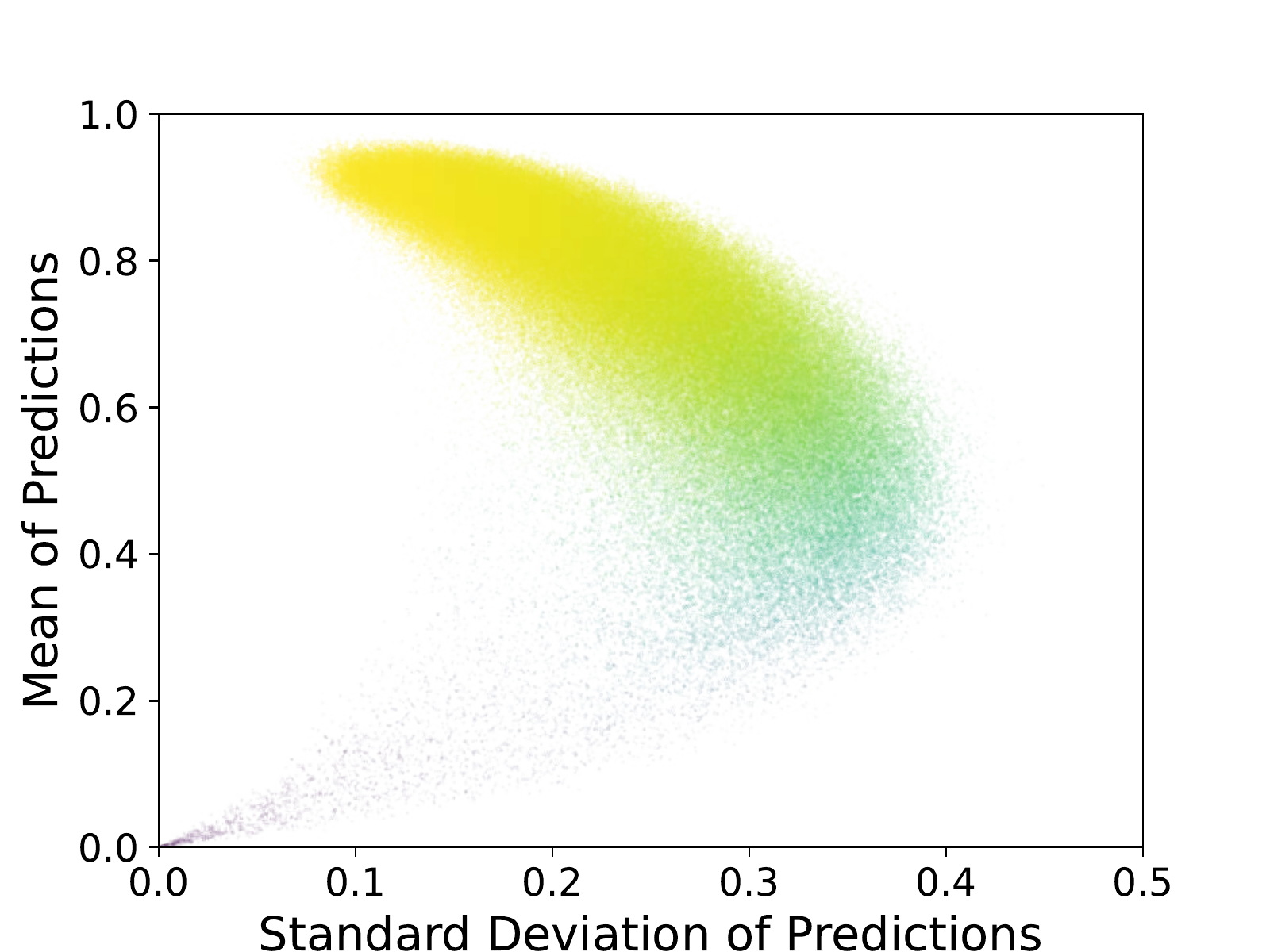}
		\subcaption{Entropy Pruned}
		\label{fig:dm_75_entropy}
	\end{minipage}
 \end{minipage}
        \begin{minipage}[c]{0.06\textwidth}
		\centering
		\includegraphics[width=\textwidth]{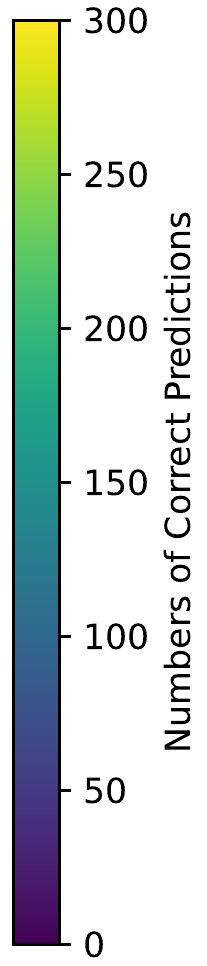}
	\end{minipage}
	\caption{Visualization of data distribution and pruned data. We train a Swin Transformer on ImageNet-1K and record the predictions. The mean and standard deviation of predictions over all epochs are calculated for each sample, i.e., each ponit in the figure. The color represents the number of correct predictions in all epochs. (a) The distribution of the whole train set of ImageNet-1K. (b) - (h) The distribution of samples pruned by every method at the pruning ratio of $25\%$.}
	\label{fig:DataMap}
\vspace{-5pt}
\end{figure*}

\subsection{Analysis of Pruned Data}
Following \cite{swayamdipta2020dataset}, we analyze the training data in terms of the prediction accuracy and variance. We train a Swin Transformer on ImageNet-1K and record the predictions of each training sample throughout all training epochs. Note that the prediction is a scalar (probability) that corresponds to the true label.
Then, the mean and standard deviation of predictions over all training epochs are calculated for each training sample, which compose the y-axis and x-axis respectively. The color reflects the number of correct predictions throughout the training. As shown in \cref{fig:DataMap}, the training set can be roughly split into three groups, namely 1) easy samples that have high mean accuracy and low deviation, 2) hard samples that have low mean accuracy and low deviation, 3) uncertain samples, namely the rest ones. 
Previous works about sample selection \cite{chang2017active, coleman2020selection, swayamdipta2020dataset}, especially active learning, suggest uncertainty sampling for minimizing the uncertainty of model prediction. 
% Given a model, these methods select training samples with largest prediction uncertainty to learn or label. 
Most of these metrics are statistic, namely only one deviation/variance is calculated on a set of predictions/variables. Instead, the proposed metric is an average over a set of deviations throughout training process.

We visualize the data ($25\%$) pruned by five state-of-the-art methods and ours in \cref{fig:DataMap}. Our method remove a large ratio of easy samples with high mean accuracy and low deviation and many hard samples with both low mean accuracy and low deviation. Note that our method is different from the naive pruning with statistic metric, for example using y-axis, namely, the mean accuracy. \emph{Forgetting} method removes more easy samples while fewer hard samples, since some hard samples may cause continuous overturning, i.e., large forgetfulness, during model training. However, such samples may be less informative for model training and meanwhile slow convergence. Error-based method \emph{EL2N} remove easy samples only. \emph{Moderate} and \emph{Self-Sup. Proto.} both remove many easy samples. For both \emph{Forgetting} and ours, the uncertain samples are rarely removed, since the two methods both consider the training dynamics. Different from \emph{Forgetting} and ours, \emph{Self-Sup. Proto.}, \emph{Moderate} and \emph{Entropy} remove many uncertain samples, as they merely take the feature distribution or final prediction into consideration.

\begin{figure*}
   
 		\centering
   \begin{minipage}[c]{0.7\textwidth}
		\centering
		\includegraphics[width=\textwidth]{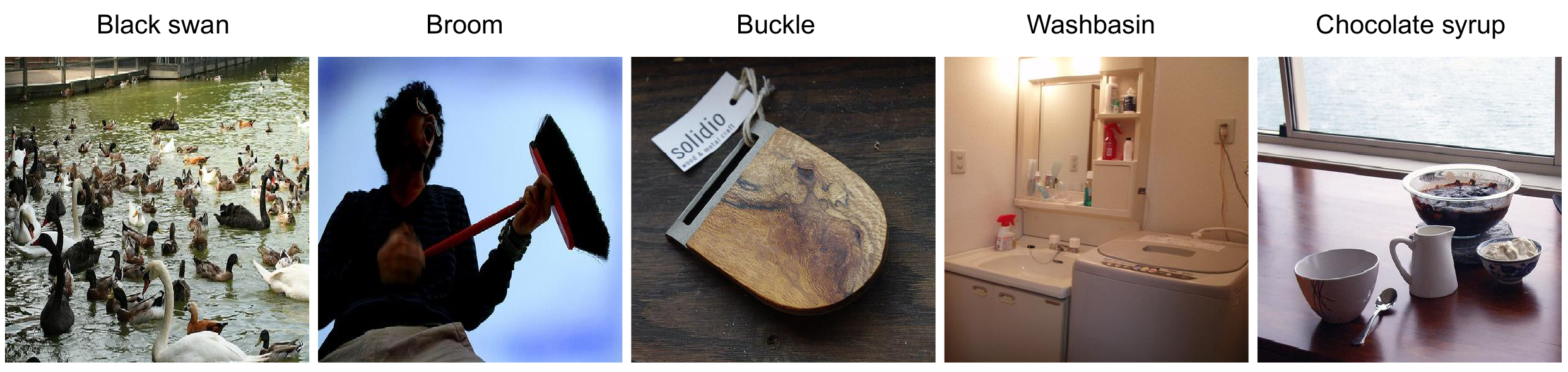}
              \subcaption{Hard samples kept by ours.}

            \label{fig:hardkeep}
	\end{minipage}\\
        \begin{minipage}[c]{0.7\textwidth}
        \begin{tabular}{c|c|c|c|c}
             % \emph{black swan, Cygnus atratus}& \emph{broom}& \emph{buckle}& \emph{stage} & \emph{toilet tissue, toilet paper, bathroom tissue}
        \end{tabular}
	\end{minipage}\\
	\begin{minipage}[c]{0.7\textwidth}
		\centering
		\includegraphics[width=\textwidth]{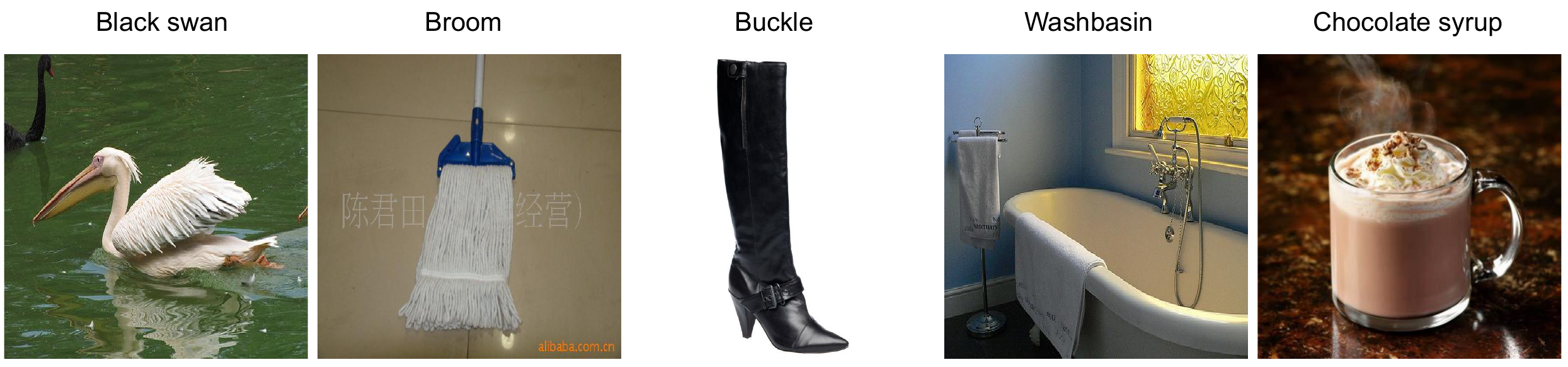}
		\subcaption{Hard samples removed by ours.}
            \label{fig:hardremove}
	\end{minipage}
	\caption{Illustrations of hard samples kept and removed by our method at the pruning ratio of $25\%$. The removed ones are likely mislabeled or confusing, while the kept ones are certainly recognizable.}
 % The class is successively \emph{black swan, Cygnus atratus}, \emph{broom}, \emph{buckle}, \emph{stage} and \emph{toilet tissue, toilet paper, bathroom tissue}. For \cref{fig:hardremove}, model predicts them as \emph{pelican}, \emph{swab, swob, mop}, \emph{cowboy boot}, \emph{theater curtain, theatre curtain} and \emph{web site, website, internet site, site} respectively with small deviation.}
	\label{fig:selected samples}

\end{figure*}

\cref{fig:selected samples} demonstrates some hard samples that are kept or removed by our method. Obviously, these hard samples removed by our method are likely mislabeled. For example, the ``cowboy boot'' is labeled as ``buckle'' but only a small region of the image contains a black buckle. The hard samples kept by our method are still recognizable thought they contain some variances. For example, the ``black swans'' are with some ``white swans'', and the ``broom'' image contains a man with a ``broom''.

% Figure~\cref{fig:DataMap}(b) shows the distribution of pruned data by our 
% variance-average score method at the ration $25\%$. Part of easy-to-learn and hard-to-learn samples are removed and most ambiguous samples are kept. The reasons are: (1) easy samples are less informative and are partly removed by almost every pruning methods;
% (2) hard samples are too hard to teach the model some knowledge and
% sometimes mislead the model due to mislabeled condition; (3) ambiguous samples contribute to the generalization performance. The following experiments prove that our pruning method is reasonable.
\section{Experiments}
\subsection{Datasets and Settings}
\paragraph{Datasets and Architectures.} We evaluate our method on ImageNet-1K and ImageNet-21K \cite{deng2009imagenet} which have $1.3$M and $14$M training samples respectively. 
We implement main experiments with Swin Transformer \cite{liu2021swin}, and also test our method on ConvNeXt \cite{liu2022convnet} and ResNet \cite{he2016deep}. 
We set the input size to be $224\times224$. 
% to explore the effectiveness under the circumstance of large-scale datasets and the new-generation backbones of computer vision.

\paragraph{Competitors.} We compare to multiple representative and state-of-the-art dataset pruning methods, namely \emph{Random}, \emph{Forgetting} \cite{toneva2018an}, \emph{Entropy} \cite{coleman2020selection}, \emph{EL2N} \cite{paul2021deep}, \emph{Self-Supervised Prototypes} \cite{sorscher2022beyond} and \emph{Moderate} \cite{xia2023moderate}. 
\cref{tab:compa_methods} summarizes the characteristics of these methods. 
Note that there exist some other pruning methods. We do not compare with them due to their difficulty in scaling up to large datasets and models \cite{yang2023dataset} or different experimental settings \cite{qin2023infobatch}. 

\begin{table*}
% \vspace{-2pt}
  \setlength{\tabcolsep}{5pt}
 \caption{Comparison to other dataset pruning methods.}
  \label{tab:compa_methods}
  \centering
  \small
  \begin{tabular}{l|cccccc}
    \toprule
    \multirow{2}{*}{Method} & Prediction &  Training & Feature & Training & Label & Class \\
     &  Uncertainty  &  Dynamics &  Distribution & Error & Supervised & Balance \\
    \midrule
    Random & & & & & &    \\
    Forgetting & \checkmark & \checkmark && & \checkmark &    \\
    Entropy &\checkmark & && &\checkmark &   \\
    EL2N & & & &\checkmark &\checkmark &   \\
    Self-Sup.Proto. & & &\checkmark& & &   \\
    Moderate & & & \checkmark& & \checkmark & \checkmark  \\
    Dyn-Unc (Ours) & \checkmark & \checkmark && &  \checkmark & \\
    \bottomrule
  \end{tabular}

\end{table*}

\paragraph{Implementation details.} All experiments are conducted with NVIDIA A100 GPUs and PyTorch \cite{paszke2019pytorch}. To accelerate data loading and pre-processing, we use NVIDIA Data Loading Library (DALI). For our method, we use $J=10$ as the window's length. All experiments are evaluated on the validation set of ImageNet-1K and ImageNet-ReaL \cite{beyer2020we}. ImageNet-ReaL reassesses the labels of ImageNet-1K-val and largely remedies the errors in the original labels, reinforcing a better benchmark. Top-1 accuracy is reported on both validation sets.

In the main experiments, we compare all seven methods with ImageNet-1K and Swin-T under the pruning ratios of $25\%$, $30\%$, $40\%$ and $50\%$. For \emph{EL2N}, the score is averaged over ten independent runs at epoch $30$. For \emph{Self-Sup. Proto.}, a Swin-B pre-trained and fine-tuned on ImageNet-1K with SimMIM \cite{xie2022simmim} serves as the feature extractor. We train a Swin-T for implementing other dataset pruning methods.
% \footnote{Provided by the authors, accessed at \url{https://drive.google.com/file/d/1xEKyfMTsdh6TfnYhk5vbw0Yz7a-viZ0w/view?usp=sharing}} 
We employ an AdamW \cite{loshchilov2018decoupled} optimizer and train models for $300$ epochs with a cosine decay learning rate scheduler and $20$ epochs of linear warm-up. The batch size is $1024$. The initial learning rate is $0.001$, and the weight decay is $0.05$. We apply AutoAugment \cite{cubuk2019autoaugment} implemented in NVIDIA DALI, label smoothing, random path drop and gradient clipping.
% Due to budget constraint, we conduct two experiments on our method and one on others.

We also implement experiments on ImageNet-21K to indicate the good generalization ability and scalability of our method. We first pre-train Swin-T on ImageNet-21K (Fall 2011 release) for $90$ epochs with $5$ epochs of warm-up. Then we fine-tune the model on ImageNet-1K for $30$ epochs with $5$ epochs of warm-up using a cosine decay learning rate scheduler, a batch size of $1024$, an initial learning rate of $4\times10^{-5}$ and a weight decay of $10^{-8}$.

\subsection{Pruning ImageNet-1K}

\cref{tab:1k_res} displays the performances of seven methods under different pruning rate.
Obviously, our method \emph{Dyn-Unc} achieves the best performances in all settings. Specifically, when pruning $25\%$ training samples, our method obtains $79.54\%$ and $84.88\%$ top-1 accuracies on ImageNet-1K-val and ImageNet-1K-ReaL respectively, which are comparable to the upper-bound performances without pruning, namely $79.58\%$ and $84.93\%$. These results indicate that our method can achieve $25\%$ lossless dataset pruning ratio. 

We observe consistent rankings of methods whether on the original validation set or ReaL.
The performance gaps between our method and others are remarkable. The gap increases for more challenging settings in which more training data are removed. For instance, our method outperforms \emph{Forgetting} by $0.84\%$, $0.87\%$, $0.94\%$ and $3.32\%$ when pruning $25\%$, $30\%$, $40\%$ and $50\%$ samples, considering that the standard deviation is around $0.2\%$.

\begin{figure*}
		\centering
	\begin{minipage}[c]{0.4\textwidth}
		\centering
		\includegraphics[width=\textwidth]{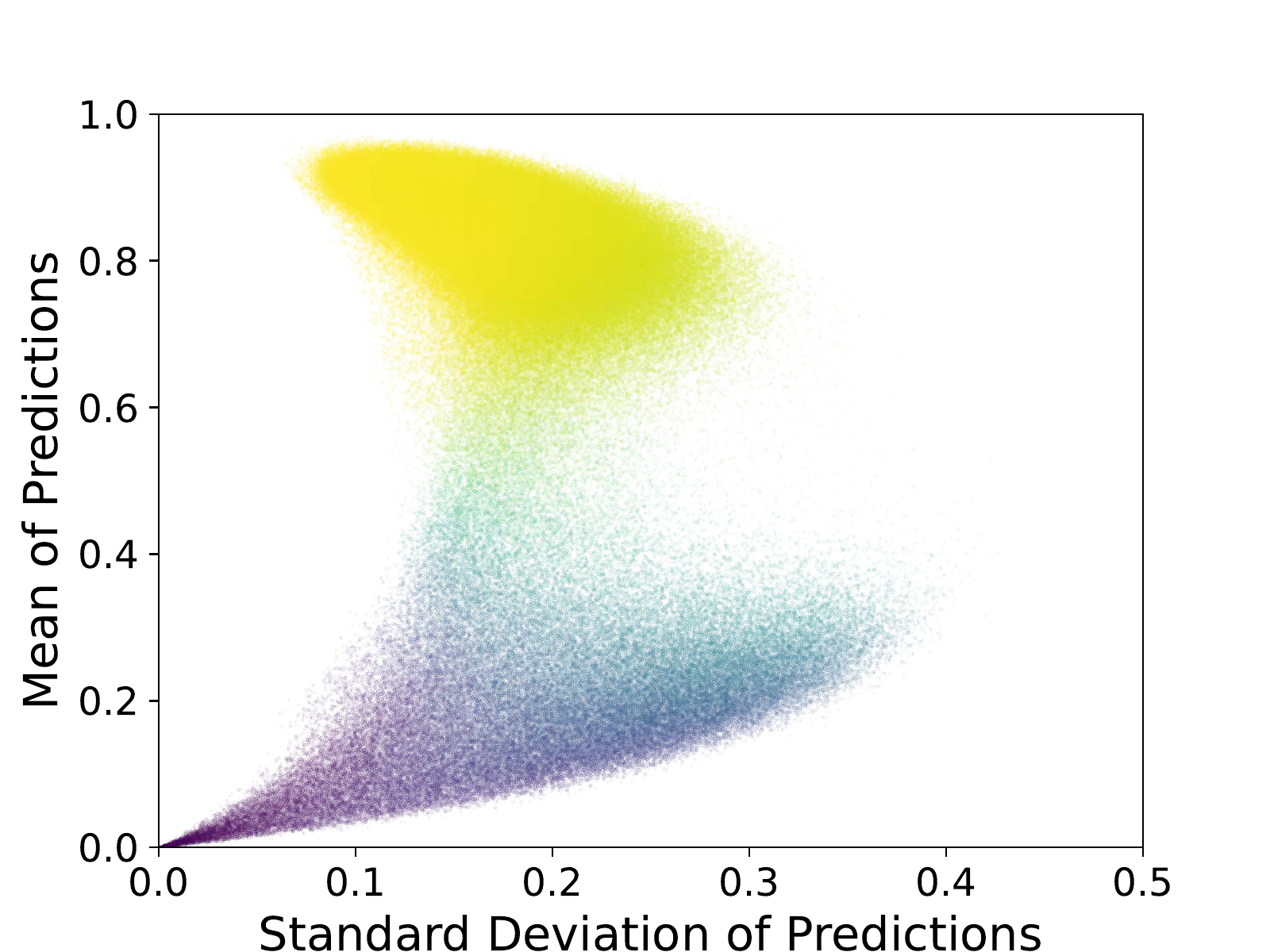}
		\subcaption{Dyn-Unc (Ours) Pruned}
		\label{fig:dm_50_ours}
	\end{minipage}
	\begin{minipage}[c]{0.4\textwidth}
		\centering
		\includegraphics[width=\textwidth]{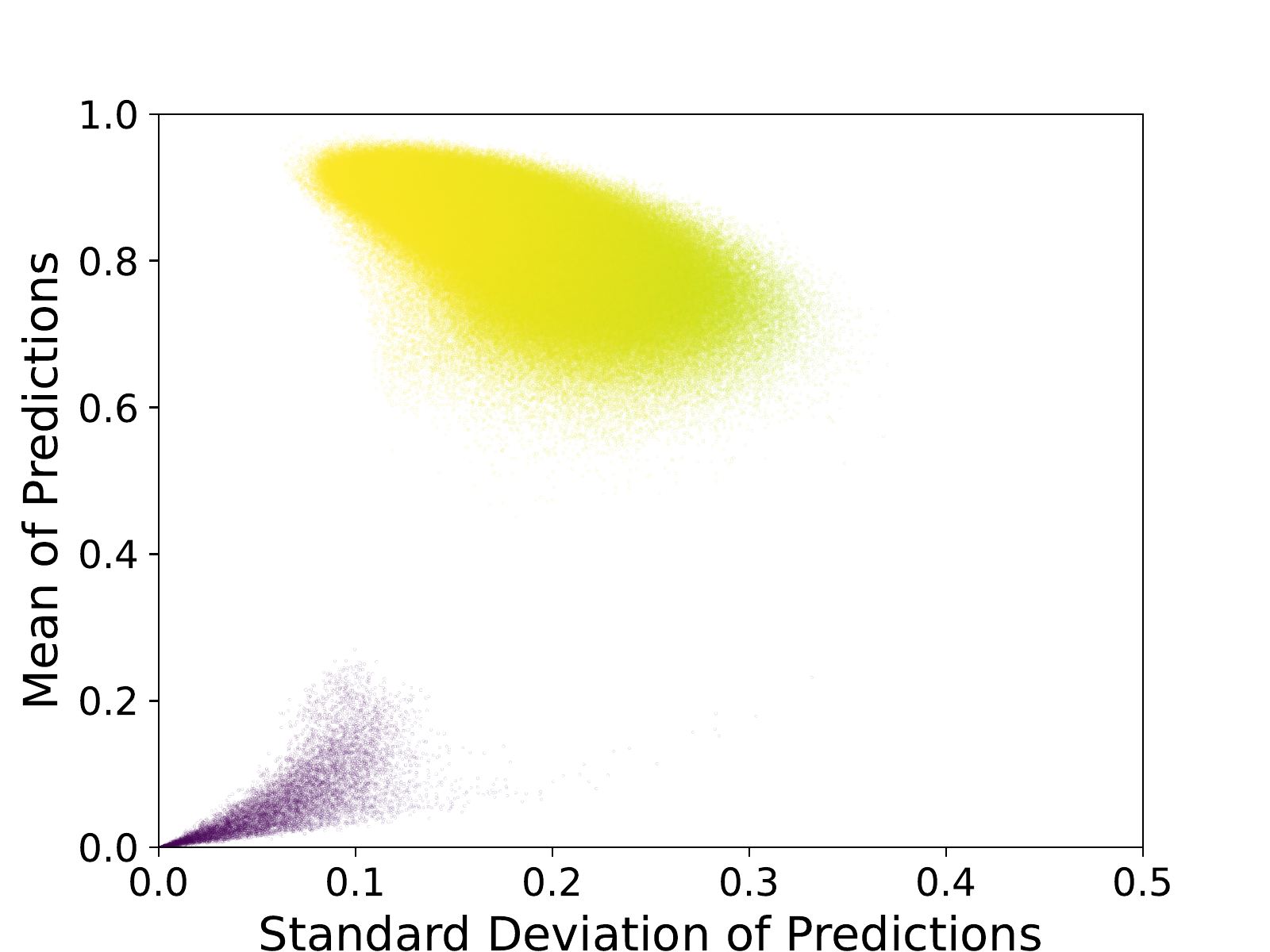}
		\subcaption{Forgetting Pruned}
		\label{fig:dm_50_forgetting}
	\end{minipage}
        \begin{minipage}[c]{0.06\textwidth}
		\centering
		\includegraphics[width=\textwidth]{figures/colorbar_v.pdf}
	\end{minipage}

	\caption{Comparison of samples pruned by ours and \emph{Forgetting} at the ratio of $50\%$. \emph{Forgetting} method throws too many easy samples.}
	\label{fig:dm_50}
\end{figure*}

\emph{Forgetting} is always the runner-up when pruning less than 40\% training data. This phenomenon emphasizes the priority of prediction uncertainty and training dynamics based methods. However, \emph{Forgetting} suffers from a large performance drop from $40\%$ to $50\%$ sample pruning. We compare the pruned samples of ours and \emph{Forgetting} in \cref{fig:dm_50}. The possible reason is that \emph{Forgetting} throws more easy samples, which makes the model training unstable. 

Generally, feature distribution based methods \emph{Self-Sup. Proto.} and \emph{Moderate} do not work well. \emph{Random} baseline exceeds all other methods except \emph{Moderate} and our \emph{Dyn-Unc} on both two validation sets when pruning $50\%$ training data. 
The average ranking shows that \emph{Entropy} performs the worst in the ImageNet-1K pruning task. The possible reason may be that this method only considers the output of
a trained model. We also illustrate these results in \cref{fig:1k_res} for easier visual comparison. 

Besides, we also test our method on ConvNeXt. With $25\%$ training data pruned, \emph{Dyn-Unc} achieves $78.77\%$ and $84.13\%$ accuracies on ImageNet-1K-val and ImageNet-1K-ReaL respectively and the results of no pruning are $78.79\%$ and $84.05\%$, suggesting that our method can generalize to different models. We further study the cross-architecture generalization ability of our method in \cref{sec:exp-cross-arc}.

% even perform worse than \emph{Random} baseline in some settings. For example, \emph{Self-Sup. Proto.} obtains worse results than \emph{Random} when pruning 40\% and  

% (1) in general, the performance rank is our method, 
% \emph{Forgetting} and the rest three, suggesting that training dynamics reveal more latent knowledge than a single training or trained model; (2)the performance of \emph{Moderate}, \emph{Self-Sup. Proto.} and \emph{Random} is pretty much the same when the pruning ratio is no more than $40\%$, and at $50\%$ \emph{Self-Sup. Proto.} drops faster; (3) there is a big drop  of \emph{Forgetting} ($3.2\%$) and \emph{Self-Sup. Proto.} ($2.1\%$) when the pruning ratio increases to $50\%$ from $40\%$, suggesting that the distribution of samples
% changes a lot at a high prune ratio under these two methods, leading to rapid declining of performance; (4) on the contrary, ours keeps a relatively stable trend when the pruned ratio gets higher but still not too high. We make a hypothesis that the sudden drop happens under all methods but at different ratio when
% the remaining samples are too hard for models to learn knowledge, as shown in Figure~\cref{fig:dm_50}.

\begin{table*}
         \vspace{-10pt}
    \caption{{Experimental results on ImageNet-1K. Our method achieves the best performances in all settings. The results indicate that our method can achieve $25\%$ lossless dataset pruning ratio.}}
    \label{tab:1k_res}
    \centering
    \small
    \resizebox{\textwidth}{!}{
        \begin{tabular}{l|cc|cc|cc|cc|c}
            \toprule
            \multicolumn{6}{l}{ImageNet-1K-val Acc. (\%)} \\
            \midrule
            Method$\backslash$Pruning Ratio & $25\%$ &rank& $30\%$& rank&$40\%$ &rank& $50\%$ &rank& avg rank \\
            \midrule
            Random & $77.82$ &5& $77.18$ &5& $75.93$ &4& $74.54$ &3& $4.25$\\
            Forgetting & $78.70$ &2& $78.27$ &2& $77.55$ &2& $74.32$ &4& $2.5$\\
            Entropy & $77.11$ &7& $76.62$  &7& $74.87$ &7& $72.65$ &6& $6.75$\\
            EL2N & $78.51$ &3& $78.15$ &3& $75.87$ &5& $71.10$ &7& $4.5$\\
            Self-Sup. Proto. & $78.24$ &4& $77.23$ &4& $75.69$ &6& $73.48$ &5& $4.75$\\
            Moderate & $77.74$ &6& $77.06$ &6& $75.94$ &3& $74.98$ &2& $4.25$\\
            Dyn-Unc (ours) & $\mathbf{79.54}\pm0.13$ &$\mathbf{1}$& $\mathbf{79.14}\pm0.07$ &$\mathbf{1}$& $\mathbf{78.49}\pm0.22$ &$\mathbf{1}$& $\mathbf{77.64}\pm0.17$ &$\mathbf{1}$& $\mathbf{1}$\\
            \midrule
            No Pruning & \multicolumn{8}{c}{$79.58\pm0.15$} & \\
        \bottomrule
        \toprule
        \multicolumn{6}{l}{ImageNet-1K-ReaL Acc. (\%)} \\
            \midrule
            Method$\backslash$Pruning Ratio & $25\%$ &rank& $30\%$& rank&$40\%$ &rank& $50\%$ &rank& avg rank \\
            \midrule
            Random & $83.55$ &5& $82.96$ &5& $81.90$ &4& $80.82$ &3& $4.25$\\
            Forgetting & $84.20$ &2& $83.75$ &2& $82.94$ &2& $79.62$ &4& $2.5$\\
            Entropy & $82.77$ &7& $82.45$ &7& $80.85$ &7& $78.96$ &6& $6.75$ \\
            EL2N & $83.84$ &3& $83.40$ &3& $81.12$ &6& $75.93$ &7& $4.75$\\
            Self-Sup. Proto. & $83.71$ &4& $82.74$ &6& $81.33$ &5& $79.28$ &5& $5$\\
            Moderate & $83.42$ &6& $83.00$ &4& $81.95$ &3& $81.05$ &2& $3.75$\\
            Dyn-Unc (ours) & $\mathbf{84.88}\pm0.15$ &$\mathbf{1}$& $\mathbf{84.62}\pm0.05$ &$\mathbf{1}$& $\mathbf{84.13}\pm0.26$ &$\mathbf{1}$& $\mathbf{83.22}\pm0.19$ &$\mathbf{1}$& $\mathbf{1}$\\
            \midrule
            No Pruning & \multicolumn{8}{c}{$84.93\pm0.11$} & \\
        \bottomrule
      \end{tabular}
    }
    \vspace{-10pt}
\end{table*}

\subsection{Pruning ImageNet-21K}

In this paper, we for the first time test dataset pruning algorithms on ImageNet-21K which consists of $14$ million images. The expensive computational cost prevents previous works from studying this dataset. As shown in \cref{tab:21k_res}, we prune ImageNet-21K with different ratios and then test on ImageNet-1K-val and ImageNet-1K-ReaL. When pruning $25\%$ training data, the Swin-T trained on our pruned dataset can achieve $82.14\%$ and $87.31\%$ accuracies on ImageNet-1K-val and ImageNet-1K-ReaL respectively. The former has $0.13\%$ performance drop while the latter has $0.07\%$ improvement. Considering the deviation, the results indicate that our method can achieve $25\%$ lossless pruning ratio on ImageNet-21K.

\begin{table}
        \caption{Experimental results on ImageNet-21K.}
        \label{tab:21k_res}
        \centering
        \small

        \begin{tabular}{c|cc}
            \toprule
            \multicolumn{3}{c}{Accuracy ($\%$)}\\
            \midrule
            Pruning Ratio& IN-1K-val & IN-1K-ReaL\\
            \midrule
            $0\%$ & $\mathbf{82.27}$  & $\mathbf{87.24}$\\
            $25\%$ & $\mathbf{82.14}$  & $\mathbf{87.31}$\\
            $30\%$ & $81.96$  & $87.16$\\
            $40\%$ & $81.68$  & $87.15$\\
            $50\%$ & $81.26$  & $86.84$\\
        \bottomrule
      \end{tabular}
      
\end{table}

% We admit that the performance loss may be larger with another downstream task, but the result shows ours' potential of scaling to huge datasets. An informative coreset is much easier to distribute and helps to more effective and efficient training, especially in the era of large models.

\subsection{Cross-architecture Generalization}
\label{sec:exp-cross-arc}
Few works test the cross-architecture generalization performance of pruned dataset, although it is important for real-world applications in which the new architectures will be trained.
We conduct experiments to verify the pruned dataset can generalize well to those unseen architectures that are inaccessible during dataset pruning. The representative architectures, Swin-T, ConvNeXt-T and ResNet-50, are taken into consideration here. Dataset pruned by one architecture is used to train others. We do experiments at the pruning ratio of $25\%$. We train ConvNeXt-T with the batch size of $2048$, initial learning rate of $0.008$. The rest hyper-parameters are the same as training Swin-T. For training ResNet-50, we employ the SGD optimizer for $90$ epochs with step decay, batch size of $2048$, initial learning rate of $0.8$ and weight decay of $10^{-4}$.

The results in \cref{tab:cross_res} show that the dataset pruned by our method has good generalization ability to unseen architectures. For example, pruning ImageNet-1K with Swin-T and then training ConvNeXt-T and ResNet-50 causes only $0.13\%$ and $0.4\%$ performance drop on ImageNet-1K-val compared to training the two architectures on the whole dataset respectively. Notably, pruning with Swin-T or ResNet-50 leads to performance improvement of training ConvNeXt-T over the whole dataset training when trained models are validated on ImageNet-1K-ReaL. It is reasonable because dataset pruning with proper ratio can remove noisy and less-informative data, thus improve dataset quality.

We further compare to \emph{Forgetting} and \emph{Moderate} in cross-architecture experiment of pruning ImageNet-1K with Swin-T and then testing with other architectures. As shown in \cref{tab:cross_res_compa}, our method performs slightly worse than the upper-bound of no pruning, but remarkably exceeds
other methods. Especially, our method surpasses \emph{Forgetting} and \emph{Moderate} by $0.95\%$ and $1.68\%$ when testing with ResNet-50 on ImageNet-1K-val. 
The promising results indicate that our method has good generalization ability on
unseen architectures and the pruned dataset can be used in real-world applications in which the architectures of downstream tasks are unknown.

\begin{table}
   \vspace{-5pt}
   \caption{Cross-arch. generalization performance of Dyn-Unc.}
    \label{tab:cross_res}
    \centering
        \small

    \begin{tabular}{l|cc}
        \toprule
        \multicolumn{3}{c}{Accuracy ($\%$)}\\
        \midrule
        Pruning $\rightarrow$ Test Model& IN-1K-val & IN-1K-ReaL\\
        \midrule
        ConvNeXt $\rightarrow$ Swin & $78.82$ & $84.05$\\
        ResNet $\rightarrow$ Swin & $79.03$  & $84.78$\\
        Swin No Pruning & $79.58$ & $84.93$\\
        \midrule
        Swin $\rightarrow$ ConvNeXt & $78.66$ & $84.15$\\
        ResNet $\rightarrow$ ConvNeXt & $78.68$  & $84.26$\\
        ConvNeXt No Pruning & $78.79$ & $84.05$\\
        \midrule
        Swin $\rightarrow$ ResNet & $75.51$ & $82.17$\\
        ConvNeXt $\rightarrow$ ResNet & $75.39$  & $82.10$\\
        ResNet No Pruning & $75.91$ & $82.64$\\
    \bottomrule
  \end{tabular}
\end{table}

\begin{table}
      \vspace{-5pt}
 \caption{Comparison to others w.r.t. cross-arch. generalization.}
    \label{tab:cross_res_compa}
    \centering
        \small

    \begin{tabular}{l|cc}
        \toprule
        \multicolumn{3}{c}{Swin $\rightarrow$ ConvNeXt Acc. (\%)} \\
        \midrule
        Method$\backslash$Val. Set& IN-1K-val & IN-1K-ReaL \\
        \midrule
        Forgetting & $78.50$  & $83.84$\\
        Moderate & $77.25$ & $82.91$ \\
        Dyn-Unc (ours) & $78.66$ & $84.15$ \\
        \midrule
        No Pruning & $78.79$ & $84.05$ \\
    \bottomrule
    \toprule
        \multicolumn{3}{c}{Swin $\rightarrow$ ResNet Acc. (\%)} \\
        \midrule
        Method$\backslash$Val. Set& IN-1K-val & IN-1K-ReaL \\
        \midrule
        Forgetting & $74.56$  & $81.52$ \\
        Moderate & $73.83$ & $80.91$\\
        Dyn-Unc (ours) & $75.51$ & $82.17$ \\
        \midrule
        No Pruning & $75.91$& $82.64$  \\
    \bottomrule
  \end{tabular}
\end{table}

\subsection{Out-of-distribution Detection}

% \begin{wraptable}{r}{0.5\textwidth}
%     \caption{Out-of-distribution detection performance of Dyn-Unc (ours).}
%     \label{tab:ood_res}
%     \centering
%     \begin{tabular}{c|c}
%         \toprule
%         \multicolumn{2}{c}{Swin-T} \\
%         \midrule
%         Pruning Ratio & ImageNet-O AUPR (\%) \\
%         \midrule
%         $0\%$ & $21.98\pm0.14$\\
%         $25\%$ & $22.51\pm0.38$\\
%         $30\%$ & $22.61\pm0.27$ \\
%         $40\%$ & $21.97\pm0.44$ \\
%         $50\%$ & $21.85\pm0.40$\\
%     \bottomrule
%   \end{tabular}
% \end{wraptable}
To further verify the robustness and reliability of our method, we evaluate our method on ImageNet-O \cite{hendrycks2021natural}, an out-of-distribution detection dataset designed for evaluating models trained on ImageNet. 
ImageNet-O consists of images from classes that are not found in ImageNet-1K but in ImageNet-21K. ImageNet-O collects images that are likely to be wrongly classified with high confidence by models trained on ImageNet-1K. Hence, this dataset can be used to test the robustness and reliability of models.
% It is highly possible for images in ImageNet-O to be assigned to wrong labels but with high confidence prediction of ImageNet-1K classes by the model due to the detailed design of ImageNet-O. 
We use AUPR (area under the precision-recall curve) metric \cite{saito2015precision,hendrycks2021natural}, and higher AUPR means better performing OOD detector. For ImageNet-O, random chance level for the AUPR is approximately $16.67\%$, and the maximum is $100\%$ \cite{hendrycks2021natural}.

Based on models trained in the aforementioned ImageNet-1K experiments, we report the OOD detection results in \cref{tab:ood_res}. Interestingly, our dataset pruning method can improve the OOD detection performance slightly, from $21.98\%$ (no pruning) to $22.61\%$ (pruning 30\% samples). The possible reason is that dataset pruning prevents models from overfitting many easy and noisy samples, and thus improves model's generalization ability.

\begin{table}
        \vspace{-5pt}
\caption{Out-of-distribution detection performance of Dyn-Unc.}
    \label{tab:ood_res}
    \centering
     \small

    \begin{tabular}{c|c}
        \toprule
        \multicolumn{2}{c}{Swin-T} \\
        \midrule
        Pruning Ratio & ImageNet-O AUPR (\%) \\
        \midrule
        $0\%$ & $21.98\pm0.14$\\
        $25\%$ & $22.51\pm0.38$\\
        $30\%$ & $22.61\pm0.27$ \\
        $40\%$ & $21.97\pm0.44$ \\
        $50\%$ & $21.85\pm0.40$\\
    \bottomrule
  \end{tabular}
\end{table}

\subsection{Ablation Study of Observation Number}
\label{sec:abl_observ}
\cite{swayamdipta2020dataset} conducts experiments on NLP tasks and shows that training on ambiguous data can achieve quite good in-distribution and out-of-distribution performance. We apply it to image classification task. Different from our method, \cite{swayamdipta2020dataset} takes first five or six epochs into account and observe only one variance of these epochs. We compare to this method in \cref{tab:ob_num_impact}. The significant performance gap implies that measuring one variance is not enough and training dynamics is important for sample selection.

Our \emph{Dyn Unc} observes $K-J$ variances throughout the $K$-epoch training with a sliding window of length $J$. 
We try to figure out whether more observations from multiple model training processes with different initializations and architectures (Swin-T, ConvNeXt and ResNet) can produce better metric and therefore better performance. 
In \cref{tab:ob_num_impact}, the results show that using more observations does not improve the performance. 
\begin{table}
    \caption{\footnotesize{Ablation study on observation number at $25\%$ pruning ratio.}}
    \label{tab:ob_num_impact}
    \centering
     \small
   \begin{tabular}{l|cc}
        \toprule
        \multicolumn{3}{c}{Accuracy ($\%$)}\\
        \midrule
        Method (observation numbers)& IN-1K-val & IN-1K-ReaL \\
        \midrule
        \cite{swayamdipta2020dataset} (only 1)  & $75.05$  & $81.60$\\ %($1$)
        Dyn-Unc ($1\times$, Ours) & $79.54$ & $84.88$ \\
        Multi. Init. ($3\times$) & $79.19$ & $84.75$ \\
        Multi. Arch. ($3\times$) & $79.34$ & $84.60$ \\
    \bottomrule
  \end{tabular}
\end{table}
\section{Conclusion}
In this paper, we push the study of dataset pruning to large datasets, i.e., ImageNet-1K/21K and advanced models, i.e., Swin Transformer and ConvNeXt. A simple yet effective dataset pruning method is proposed based on the prediction uncertainty and training dynamics. The extensive experiments verify that our method achieves the best results in all settings comparing to the state of the art. Notably, our method achieves $25\%$ lossless pruninng ratio on both ImageNet-1K and ImageNet-21K. The cross-architecture generalization and out-of-distribution detection experiments show promising results that pave the way for real-world applications.

\paragraph{Acknowledgement.}
This work is funded by National Science and Technology Major Project (2021ZD0111102) and NSFC-62306046.

{
    \small
    \bibliographystyle{ieeenat_fullname}
    \bibliography{main}
}


\end{document}